%% file: main.tex
\documentclass[sigconf]{acmart}
\usepackage{flushend}
\usepackage{caption}
\usepackage{tabularx}
\usepackage{graphicx}
\usepackage{amsfonts}
\usepackage{amsmath}
\usepackage{amsthm}
\usepackage{algorithm}
\usepackage{algorithmic}
\usepackage{multirow}
\usepackage{booktabs}
\usepackage{float}
\usepackage{bm}
\usepackage{xspace}
\usepackage{threeparttable}
\usepackage{hyperref}
\usepackage{url}
\usepackage{mathtools}
\usepackage{makecell}
\usepackage{subcaption}
\usepackage{color}
\usepackage{xcolor}
\usepackage{balance}
\usepackage{enumitem}

\usepackage{amssymb}

\newcommand{\eg}{\textit{e}.\textit{g}.}

\newtheorem*{Pro*}{Problem}
\newcommand{\model}{ScaleGNN\xspace}
\AtBeginDocument{%
  }

\setcopyright{acmlicensed}

\copyrightyear{2026}
\acmYear{2026}
\setcopyright{cc}
\setcctype{by}
\acmConference[WWW '26]{Proceedings of the ACM Web Conference 2026}{April 13--17, 2026}{Dubai, United Arab Emirates}
\acmBooktitle{Proceedings of the ACM Web Conference 2026 (WWW '26), April 13--17, 2026, Dubai, United Arab Emirates}
\acmPrice{}
\acmDOI{10.1145/3774904.3792347}
\acmISBN{979-8-4007-2307-0/2026/04}

\begin{document}

\title{ScaleGNN: Towards Scalable Graph Neural
Networks via Adaptive High-order Neighboring
Feature Fusion}

\author{Xiang Li}
\email{lixiang1202@stu.ouc.edu.cn}
\affiliation{%
  \institution{Ocean University of China}
  \city{Qingdao}
  \country{China}
}

\author{Jianpeng Qi}
\email{qijianpeng@ouc.edu.cn}
\affiliation{%
  \institution{Ocean University of China}
  \city{Qingdao}
  \country{China}
}

\author{Haobing Liu}
\email{haobingliu@ouc.edu.cn}
\affiliation{%
  \institution{Ocean University of China}
  \city{Qingdao}
  \country{China}
}

\author{Yuan Cao}
\email{cy8661@ouc.edu.cn}
\affiliation{%
  \institution{Ocean University of China}
  \city{Qingdao}
  \country{China}
}

\author{Guoqing Chao}
\email{guoqingchao@hit.edu.cn}
\affiliation{%
 \institution{Harbin Institute of Technology}
 \city{Weihai}
 \country{China}
}

\author{Zhongying Zhao}
\email{zyzhao@sdust.edu.cn}
\affiliation{%
  \institution{Shandong University of Science and Technology}
  \city{Qingdao}
  \country{China}
}

\author{Junyu Dong}
\email{dongjunyu@ouc.edu.cn}
\affiliation{%
  \institution{Ocean University of China}
  \city{Qingdao}
  \country{China}
}

\author{Xinwang Liu}
\email{xinwangliu@nudt.edu.cn}
\affiliation{%
  \institution{School of Computer, National University of Defense Technology}
  \city{Changsha}
  \country{China}
}

\author{Yanwei Yu}
\authornote{Yanwei Yu is the corresponding author.}
\email{yuyanwei@ouc.edu.cn}
\affiliation{%
  \institution{Ocean University of China}
  \city{Qingdao}
  \country{China}
}

\renewcommand{\shortauthors}{Xiang Li et al.}

\begin{abstract}
Graph Neural Networks (GNNs) have demonstrated impressive performance across diverse graph-based tasks by leveraging message passing to capture complex node relationships. However, on large-scale real-world graphs, GNNs face two major challenges: (1) GNNs struggle to ensure scalability and efficiency as repeated aggregation of large neighborhoods incurs significant computational overhead; (2) GNNs suffer from over-smoothing, where excessive propagation makes node representations indistinguishable, hindering model expressiveness. To tackle these, we propose ScaleGNN, which adaptively fuses multi-hop node features for scalable and effective graph learning. We first compute per-hop pure-neighbor matrices to isolate exclusive structural signals, then apply lightweight fusion to balance low- and high-order information, preserving both local detail and global correlations. To curb redundancy and over-smoothing, we introduce Local Contribution Score (LCS)–based masking to prune low-relevance high-order neighbors, and impose learnable sparsity to selectively integrate valuable multi-hop features. Extensive experiments on real-world datasets show that ScaleGNN consistently outperforms state-of-the-art GNNs in both predictive accuracy and computational efficiency. The source code is available at \url{https://github.com/lx970414/ScaleGNN}. 
\end{abstract}




\maketitle

\input{Main/1_Introduction}
\input{Main/2_Related_Work}

\input{Main/3_Preliminary}
\input{Main/4_Methodology}

\input{Main/5_Experiment}

\input{Main/6_Conclusion}

\begin{acks}
This work is supported by the Advanced Materials-National Science and Technology Major Project (Grant No. 2024ZD0607700).
\end{acks}

\clearpage
\bibliographystyle{ACM-Reference-Format}
\bibliography{www26}

\appendix

\input{Main/7_Appendix}

\end{document}

%% file: Main/1_Introduction.tex
\section{Introduction}
Graph Neural Networks (GNNs) have become the cornerstone of modern graph representation learning~\cite{hou2023graphmae2,li2024long,liang2025multi,tu2025wage}, driving advances in applications such as node classification~\cite{hu2024efficient}, link prediction~\cite{yu2022multiplex,hang2024paths2pair}, anomaly detection~\cite{zeng2024mitigating,li2025umgad,zeng2025imol,zeng2025understand}, and recommender systems~\cite{li2023zebra,shui2024hierarchical,li2025dual}. The essence of GNNs lies in iterative message-passing schemes~\cite{wu2022graph,fu2023multiplex}, whereby nodes update their representations by aggregating information from local neighborhoods~\cite{zheng2023node}. This capability allows GNNs to naturally encode node representations and capture the underlying correlations between nodes.

As digital infrastructures evolve, graphs with millions or even billions of nodes and edges have become commonplace, magnifying the necessity for GNNs that are not only expressive but also computationally scalable~\cite{zhang2022oag,zheng2023structure,kong2023goat}. In response, recent research communities have developed a rich array of scalable and deep GNN architectures, with significant innovations focused on balancing model expressiveness, computational efficiency, and training feasibility. \textit{One line of research is sampling-based scalable GNN methods} such as GraphSAGE~\cite{hamilton2017inductive} and Cluster-GCN~\cite{chiang2019cluster}, which attempt to control computational costs by selecting subsets of neighbors during training, thereby reducing memory overhead and enabling mini-batch processing. These approaches help alleviate the burden of large-scale aggregation but may lose important structural information due to incomplete sampling. \textit{Another stream is pre-computation-based scalable approaches} (including SGC~\cite{wu2019simplifying}, SIGN~\cite{rossi2020sign}, and GAMLP~\cite{zhang2022graph}), which decouple neighborhood aggregation from parameter learning, allowing feature propagation to be computed in advance and greatly accelerating both training and inference. \textit{Meanwhile, a growing body of research has focused on deep and high-order GNNs} (\eg, S\textsuperscript{2}GC~\cite{zhu2021simple}, GBP~\cite{chen2020scalable}, RpHGNN~\cite{finkelshtein2024learning}). They integrate information from high-order neighbors, thereby enriching the receptive field of each node and improving the capture of long-range dependencies. These advances have substantially expanded the applicability of GNNs to web-scale scenarios such as large-scale heterogeneous graphs~\cite{zhang2022oag,zheng2023structure,kong2023goat}.

Despite these substantial developments, GNNs still face two persistent and intertwined challenges when deployed on large-scale and complex graphs. \textit{First, as the number of message-passing layers increases, GNNs suffer from the over-smoothing problem that node representations become overly similar and lose their discriminative power}. This issue is particularly severe when incorporating high-order neighbors~\cite{jin2022graph,zhang2024hongat,gong2024hn}, as their contributions may become redundant or even introduce noise, leading to performance degradation. Although several techniques, such as residual connections~\cite{li2021training}, skip connections~\cite{wu2019simplifying,chen2022bag}, and decoupling propagation~\cite{zeng2021decoupling} from feature transformation, have been proposed to alleviate over-smoothing, they often introduce additional computational complexity or fail to effectively balance local and global information. \textit{Second, many traditional GNN architectures struggle with the scalability issue that models face high consumption and computational costs when applied to large-scale graphs}~\cite{zhang2018billion,liu2020fast,li2021higher,ding2022sketch,chen2024macro}. The primary reason is the exponential growth in the number of high-order neighbors, which leads to excessive information aggregation and redundant computations. Some recent works address scalability issues~\cite{finkelshtein2024learning} by adopting pre-computation-based techniques, mini-batch training, or sampling strategies. While these approaches improve efficiency, they often sacrifice performance due to incomplete multi-hop neighborhood information. Recently, plug-in modules such as RMask~\cite{liang2025towards} have been introduced to further enhance scalable GNNs. These modules can be seamlessly integrated into frameworks like S\textsuperscript{2}GC and GAMLP, and employ techniques such as noise masking or random walk-based neighbor selection to reduce redundancy and over-smoothing. While plug-in enhancements offer practical improvements, most still fundamentally inherit the aggregation schemes of their base models~\cite{jin2022graph,zhang2024hongat,gong2024hn} and often lack adaptive, effective feature fusion of information from different neighborhood hops. \textit{Overall}, despite advances in sampling-based, pre-computation-based, deep and high-order aggregation, and plug-in modules, scalability and over-smoothing issues still hinder scalable GNNs for large-scale and complex graphs. 
To overcome these challenges, we propose \textit{\textbf{ScaleGNN}}, a scalable framework that mitigates over-smoothing via adaptive fusion of multi-hop graph features. Specifically, we introduce a learnable mechanism to construct and refine per-hop neighbor matrices, with trainable weights adjusting the relative importance of different neighborhood orders to emphasize informative high-order neighbors while down-weighting less useful ones. A low-order–enhanced fusion further integrates low- and high-order features based on task relevance, enabling \model to capture both local structure and global correlations with limited complexity. Furthermore, we employ a selective neighbor masking strategy, the Local Contribution Score (LCS), and impose LCS-based sparsity to prune low-relevance high-order neighbors, which quantifies their contribution to target nodes and curbs redundancy and over-smoothing. Extensive experiments on multiple real-world graphs show that \model consistently outperforms state-of-the-art GNNs in both accuracy and efficiency, providing a robust solution to scalability and over-smoothing in large, complex graphs.

The main contributions of our work are as follows:
\begin{itemize}[leftmargin=*]
\item We design a trainable mechanism to construct and refine multi-hop neighbor matrices, effectively retaining informative high-order neighbors while filtering out redundancy and noise for better high-order information integration. 
\item We propose a low-order enhanced fusion strategy that adaptively weights low-order and high-order features, enabling the model to capture both local detail and global correlations while alleviating over-smoothing. 
\item We introduce the selective masking strategy named LCS to reduce multi-hop redundancy and impose sparse constraints based on LCS to prune unimportant high-order neighbors, significantly reducing computational cost with minimal performance loss. 
\item Experiments on real-world graphs show that our model consistently outperforms state-of-the-art GNNs both in accuracy and efficiency, demonstrating superior scalability. 
\end{itemize}

%% file: Main/2_Related_Work.tex
\section{Related Work}
\subsection{Graph Neural Network}
GNNs learn on graphs via message passing. Representative homogeneous GNNs include GCN~\cite{kipf2016semi}, GAT~\cite{velivckovic2017graph}, GraphSAGE~\cite{hamilton2017inductive}, SGC~\cite{wu2019simplifying}, APPNP~\cite{gasteiger2018predict}, and S\textsuperscript{2}GC~\cite{zhu2021simple}, which differ mainly in neighborhood aggregation and propagation strategies. Despite strong performance, these models often degrade on real-world graphs with heterophily~\cite{pan2023beyond,li2024pc} or heterogeneity. We focus on the latter: by assuming homogeneous structures, these methods fail to capture type-specific semantics in heterogeneous graphs~\cite{du2025contrastive}, where nodes and edges contribute unequally. Heterogeneous GNNs (HGNNs) are thus proposed to explicitly model such typed interactions.

Existing HGNNs can be broadly categorized into relation-wise and representation-wise approaches. Relation-wise methods aggregate neighbors along relations or meta-paths, such as HAN~\cite{wang2019heterogeneous}, MAGNN~\cite{fu2020magnn}, and HetGNN~\cite{zhang2019heterogeneous}. Representation-wise methods perform message passing with type- or relation-aware parameters, including R-GCN~\cite{schlichtkrull2018modeling}, RSHN~\cite{zhu2019relation}, and Transformer-based models such as HGT~\cite{hu2020heterogeneous}, Simple-HGN~\cite{lv2021we}, and HINormer~\cite{mao2023hinormer}. Recent work HGAMLP~\cite{liang2024hgamlp} improves scalability by adaptively fusing local and global information on large heterogeneous graphs. \textit{Nevertheless, most HGNNs still rely on extensive message passing, leading to high training costs on large-scale graphs.}

\begin{figure*}[t]\label{framework}
    \begin{center}
    \vspace{-4mm}
    \includegraphics[width=1\textwidth]{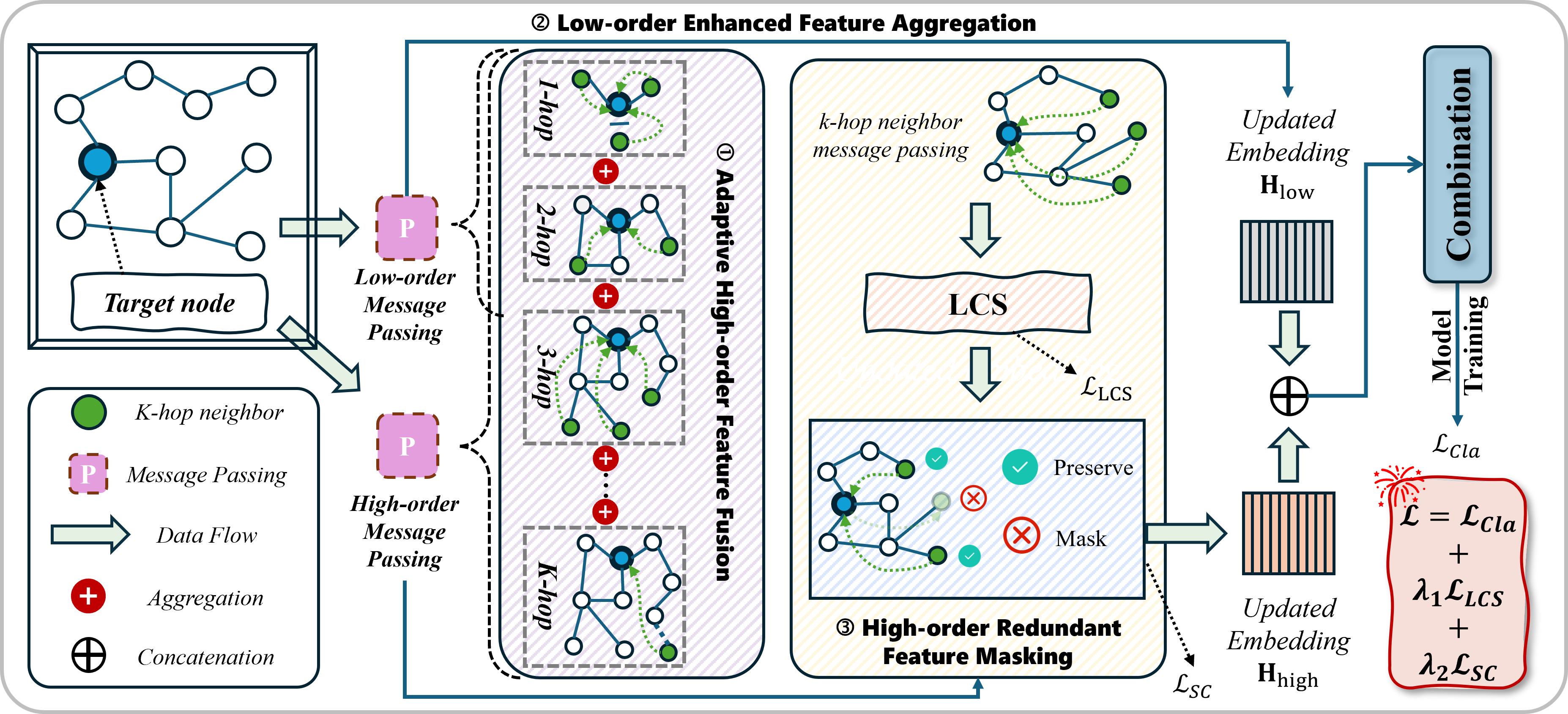}
    \vspace{-4mm}
    \caption{The overview of the proposed ScaleGNN. To address the scalability problem, ScaleGNN employs a single convolution process with a hybrid adjacency matrix that fuses features from all different neighbor hops, significantly improving computational efficiency. To mitigate the over-smoothing issue, we introduce a high-order redundant feature masking strategy, selectively filtering out less relevant high-order neighbors, which helps alleviate the over-smoothing issue.}
    \label{fig:model}
    \end{center}
    \vspace{-4mm}
\end{figure*}

\subsection{Deep Graph Neural Networks}
Deep GNNs~\cite{li2021training,zeng2021decoupling,giraldo2023trade,peng2024beyond,ju2024comprehensive} have emerged as a prevailing paradigm for scalable graph representation learning~\cite{chen2022bag,lee2023towards,zhang2024hongat}, enabling effective modeling of complex dependencies in large graphs. By stacking multiple layers, they iteratively aggregate neighborhood information to capture both local and global structures. Deep GNNs can be broadly categorized into spectral and spatial methods. Spectral approaches operate in the graph spectral domain, while spatial methods aggregate directly on graph neighborhoods and are more suitable for inductive learning. Representative spatial models include GraphSAGE~\cite{hamilton2017inductive} and GAT, with later variants such as GCNII and GATv2 further improving depth, aggregation, and expressivity.

Despite these advances, increasing depth often leads to over-smoothing and rising computational cost on large-scale or dense graphs~\cite{nie2024fast}. Techniques such as residual connections, identity mapping, decoupled propagation, and attention-based neighbor selection alleviate these issues but remain insufficient. \textit{Consequently, challenges in scalability, over-smoothing, and effective utilization of high-order information persist, motivating more adaptive and efficient deep GNN frameworks.}

\subsection{Scalable Graph Neural Networks}
To scale GNNs on large graphs, two main paradigms have emerged: sampling-based~\cite{ying2018graph,jiang2021pre,liao2022scara} and pre-computation-based methods~\cite{yang2023simple,hu2024efficient}. Sampling reduces computational cost by selecting subsets of neighbors but may introduce variance and information loss. In contrast, pre-computation decouples propagation from nonlinear transformation, enabling parameter-free diffusion to be computed offline and yielding more stable and efficient training. SGC~\cite{wu2019simplifying} exemplifies this idea by removing inter-layer nonlinearities, while S\textsuperscript{2}GC~\cite{zhu2021simple} and related works extend expressiveness via multi-hop propagation and adaptive fusion without sacrificing scalability.

Extending pre-computation to heterogeneous graphs is more challenging due to multiple node and edge types. NARS~\cite{yu2020scalable} samples relation subsets and performs relation-specific pre-computation to reduce information loss. SeHGNN~\cite{yang2023simple} aggregates relation-wise features within $K$ hops, preserving fine-grained signals but incurring high cost as the number of relations or $K$ increases. RpHGNN~\cite{hu2024efficient} addresses this issue with a hybrid design that balances relation-wise fidelity and representation-wise efficiency.

Recent plug-in modules, such as RMask~\cite{liang2025towards}, further enhance scalable GNNs by filtering propagation noise, mitigating over-smoothing, and reducing pre-processing overhead, enabling deeper diffusion and more effective feature aggregation on large graphs.

In summary, scalable GNNs improve efficiency through sampling, pre-computation, and plug-in designs, while relation-aware and adaptive aggregation are crucial. \textit{Nevertheless, balancing scalability and information preservation, reducing sampling variance, and alleviating over-smoothing remain open challenges.}

%% file: Main/3_Preliminary.tex
\section{Preliminary}
Generally, we consider an undirected large-scale graph as $\mathcal{G} = \{\mathcal{V}, \mathcal{E}\, \mathcal{X}\}$, where $\mathcal{V}$ is the collection of nodes, $\mathcal{E}$ is the collection of edges between nodes, and $\mathcal{X} \in \mathbb{R}^{|\mathcal{V}| \times f}$ is the collection of node attributes, where $f$ is the size of the node feature vector. 
$\mathbf{A} \in \mathbb{R}^{|\mathcal{V}| \times |\mathcal{V}|}$ denotes the adjacency matrix of graph $\mathcal{G}$. $\mathbf{D}=diag(d_1, d_2, \ldots, d_{|\mathcal{V}|}) \in \mathbb{R}^{|\mathcal{V}| \times |\mathcal{V}|}$ represents the degree matrix of $\mathbf{A}$, where $d_i=\sum\nolimits_{{v_j} \in \mathcal{V}} {\mathbf{A}_{ij}}$ denotes the degree of node $v_i$. Given the labeled node set $\mathcal{V}_l$, our goal is to predict the class labels for nodes in the unlabeled set $\mathcal{V}_u=\mathcal{V}-\mathcal{V}_l$ under the supervision of $\mathcal{V}_l$.

%% file: Main/4_Methodology.tex
\section{Methodology}
In this section, we present \model, an efficient end-to-end framework for scalable and robust GNN learning. \model addresses over-smoothing, computational inefficiency, and redundant high-order information through three synergistic components: (1) \textit{Adaptive High-order Feature Fusion}, (2) \textit{Low-order Enhanced Feature Aggregation}, and (3) \textit{High-order Redundant Feature Masking}. These modules enable \model to process large-scale graphs efficiently while preserving essential information across neighborhood hops. The overall framework is illustrated in Fig.~\ref{fig:model}.


\subsection{Adaptive High-order Feature Fusion}
Traditional GNNs are based on message passing through fixed adjacency matrices, where the adjacency structure remains constant throughout the layers. This mechanism can lead to over-smoothing, where node representations converge and lose their distinctiveness as the network depth increases. Furthermore, as nodes aggregate information from higher-order neighbors, the feature similarity across nodes can become excessive, thereby reducing the model's expressiveness.

To mitigate these issues, we propose a method for constructing distinct adjacency matrices for each neighborhood hop. This allows us to isolate the relations specific to each hop, ensuring that each neighborhood level contributes uniquely to the node representation. We define the \textit{pure} $i$-th hop adjacency matrix $\mathbf{A}_i$ as the difference between the $(i-1)$-th and $i$-th hops of adjacency matrices:
\begin{equation}
    \mathbf{A}_i = \mathbf{A}^i - \mathbf{A}^{i-1}, \quad i = 1,2,3, \dots, K,
\end{equation}
where $\mathbf{A}^i \in \mathbb{R}^{|\mathcal{V}|\times|\mathcal{V}|}$ is the $i$-th hop adjacency matrix, which captures the connections up to $i$ hops, $\mathbf{A}_i$ isolates the new relations emerging specifically at the $i$-th hop. $K$ is the maximum number of hops, also referred to as the order. Specifically, both $\mathbf{A}_1$ and $\mathbf{A}^1$ are the adjacency matrix of the graph, and $\mathbf{A}^0$ is the zero matrix. 

Incorporating information from multiple hops can provide a more comprehensive view of the graph structure. However, the challenge lies in integrating these multiple hops in a way that preserves their distinct contributions. To address this, we introduce a learnable weight set $\boldsymbol{\alpha} = [\alpha_1, \alpha_2, \dots, \alpha_K]$, where each $\alpha_i \in \mathbb{R}$ ($i \in \{1,2,\ldots,K\}$) represents the specific important variant of the $i$-th hop adjacency matrix in the final aggregation:
\begin{equation}\label{eq2}
    \tilde{\mathbf{A}} = \sum_{i=1}^{K} \alpha_i \mathbf{A}_i, \quad \sum_{i=1}^{K} \alpha_i = 1.
\end{equation}

These weights are learned through back-propagation, ensuring that the model dynamically adjusts the contribution of each neighborhood hop. The use of the softmax function to optimize these weights ensures that contributions across different hops remain balanced. Finally, we obtain the adaptive high-order fused embedding matrix $\mathbf{H} \in \mathbb{R}^{|\mathcal{V}| \times d_f}$ as follows:
\begin{equation}
    \mathbf{H} = \sigma(\tilde{\mathbf{A}} \cdot \mathcal{X} \cdot \mathbf{W}),
\end{equation}
where $d_f$ is the hidden embedding dimension, $\mathcal{X} \in \mathbb{R}^{|\mathcal{V}| \times f}$ is the input feature matrix, and $\mathbf{W} \in \mathbb{R}^{f \times d_f}$ is the learnable weight matrix for high-order features. The activation function $\sigma(\cdot)$ introduces non-linearity to the feature transformation.

\textbf{\textit{Specifically}}, we use the current model as our base version named \textbf{ScaleGNN\textsubscript{b}}, which is relatively simple and efficient. 

\subsection{Low-order Enhanced Feature Aggregation}

The integration of low-order (local) and high-order (global) features in many GNN-based models is done in a single and often static step. However, this can lead to high-order features dominating over low-order ones, which might diminish the model's ability to preserve fine-grained local information. To address this, we propose explicitly separating low-order and high-order features before fusion. First, we obtain the low-order feature matrix $\mathbf{H}_{\text{low}}$ and the high-order feature matrix $\mathbf{H}_{\text{high}}$ as follows:
\begin{equation}
    \mathbf{H}_{\text{low}} = \sigma(\mathbf{A}^{2} \cdot \mathcal{X} \cdot \mathbf{W}_{\text{low}}), \quad \mathbf{H}_{\text{high}} = \sigma(\tilde{\mathbf{A}} \cdot \mathcal{X} \cdot \mathbf{W}_{\text{high}}),
\end{equation}
where $\mathbf{W}_{\text{low}},\mathbf{W}_{\text{high}} \in \mathbb{R}^{f\times d}$ are both learnable weight matrices for low-order and high-order features, respectively. 

To ensure a balanced integration of low-order and high-order features, we introduce a learnable balancing factor $\beta$, which allows for dynamic adjustment of diverse contributions from two feature types. The final node representation $\mathbf{H}$ is the weighted sum of low-order and high-order features:
\begin{equation}
    \mathbf{H} = \beta \cdot \mathbf{H}_{\text{low}} + (1 - \beta) \cdot \mathbf{H}_{\text{high}},
\end{equation}
where $\beta$ is the hyperparameter, ensuring the model can effectively balance the trade-off between local and global information, depending on the specific task and graph structure.

This separation and fusion process enhances the model's ability to capture both fine-grained, local details and broader, global patterns in the graph, ultimately leading to more expressive and robust node embeddings.


\subsection{High-order Redundant Feature Masking}

High-order neighbors introduce diverse, yet potentially redundant, information to GNNs. These neighbors will introduce noise without an effective masking mechanism, diluting the informative signal and impairing the model’s ability to generate useful node embeddings. To address this, we propose a novel method for selecting the most relevant and masking other redundant high-order neighbors based on their local contributions to node representations.

Inspired by the scaled dot-product attention mechanism~\cite{vaswani2017attention}, we introduce the Local Contribution Score (LCS) to assign importance scores to high-order neighbors. The LCS is designed to quantify the relevance of each neighboring node of the target node by considering both structural similarity and feature alignment. Specifically, for a node $v$ and its $i$-th hop neighbor $u$, the LCS value is:
\begin{equation}\label{eq6}
    \text{LCS}(v,u,i) = \frac{ \exp \left( \frac{ (\mathbf{W}_1 \mathbf{x}_v)^\top (\mathbf{W}_2 \mathbf{x}_u) }{ \sqrt{d_f} } \right) }{ \sum\limits_{u' \in \mathcal{N}_i(v)} \exp \left( \frac{ (\mathbf{W}_1 \mathbf{x}_v)^\top (\mathbf{W}_2 \mathbf{x}_{u'}) }{ \sqrt{d_f} } \right) }.
\end{equation}

Here, $\mathbf{x}_v, \mathbf{x}_u \in \mathbb{R}^{f}$ are the input feature vectors of the target node $v$ and its $k$-hop neighbor $u$. $\mathbf{W}_1, \mathbf{W}_2 \in \mathbb{R}^{d_f \times f}$ are learnable projection matrices that map features into a latent attention space of dimension $d_f$, which is a tunable hyperparameter, and typically $d_f \le f$ to ensure efficiency. $\mathcal{N}_i(v)$ denotes the $i$-hop neighborhood of node $v$ induced by $\mathbf{A}_i$. The denominator normalizes attention scores within each local $i$-hop neighborhood using the softmax function. This equation allows the model to learn context-aware neighbor importance scores in a differentiable and robust way.

\input{Tables/Datasets}
The LCS allows us to mask redundant high-order neighbors and identify the most relevant neighbors by considering both local structure and feature similarity, enabling the model to selectively pass information from the most meaningful high-order neighbors. Once the LCS is computed for each neighbor, we select the top-$m_i$ neighbors with the highest LCS values for each node among its $i$-hop neighbors:
\begin{equation}
    \mathbf{A}_i^{\text{filter}} = \text{Top}_{m_i}(\mathbf{A}_i, \text{LCS}),
\end{equation}
where Top$_{m_i}(\cdot,\cdot)$ retains the $m_i$ entries with highest scores for each row in $\mathbf{A}_i$. Each $m_i$ specifies the number of neighbors to be sampled from the $i$-hop neighborhood of nodes. Notice that in our framework, $m_i$ ($i \in \{1,2,\dots,K\}$ is a learnable parameter. By learning distinct $m_i$ values, the model adaptively controls the neighbor sampling count at different hops, facilitating flexible and hierarchical aggregation of neighborhood information. This mechanism ensures that only the most informative high-order neighbors contribute to the final node representation, thereby effectively mitigating noise and enhancing embedding stability.  

\subsection{Joint Optimization}
To ensure the effectiveness of \model, we design an optimization framework that consists of three key loss components: the node classification task loss, the LCS regularization loss, and the sparse constraint loss on the high-order information. These losses work together to enhance model performance, suppress noise from uninformative high-order neighbors, and maintain computational efficiency. For node classification, we employ the standard cross-entropy loss as the primary task objective:
\begin{equation}
\mathcal{L}_{Cla} = - \sum_{i \in \mathcal{V}} y_i \log \hat{y}_i,
\end{equation}
where $y_i$ represents the one-hot vector of the ground-truth class label for node $i$, and $\hat{y}_i$ is the predicted probability distribution over classes generated by a softmax layer.

To further mitigate the over-smoothing issue, we impose a sparse constraint (SC) on high-order feature propagation:
\begin{equation}
\mathcal{L}_{SC} = \sum_{i=2}^{K} \| \bar{\mathbf{A}}_i^{\text{filter}} \|_1,
\end{equation}
where $\bar{\mathbf{A}}_i^{\text{filter}}$ is the filtered high-order adjacency matrix, $\bar{\mathbf{A}}_i^{\text{filter}} = \mathbb{I}\left[\mathbf{A}_i^{\text{filter}} > 0\right]$, where $\mathbb{I}[\cdot]$ denotes the element-wise indicator function, and $\| \cdot \|_1$ denotes the $\mathcal{L}_1$ normalization. This term constrains the model to minimize the number of neighbors retained for each node at each hop. 

To guide the model in selecting meaningful high-order neighbors, we also introduce an LCS-based regularization:
\begin{equation}
\mathcal{L}_{LCS} = \sum_{v \in \mathcal{V}} \sum_{i=2}^{K} \sum_{u \in \mathcal{N}_i^{\text{select}}(v)} \left( 1 - \text{LCS}(v, u, i) \right)^2,
\end{equation}
where $\mathcal{N}_i^{\text{select}}(v)$ is the retained neighboring nodes which are not been filtered. This term enforces high-order neighbors to be selected based on their LCS values, reducing the influence of noisy or redundant neighbors. The final optimization objective is as follows:
\begin{equation}
\mathcal{L} = \mathcal{L}_{Cla} + \lambda_1 \mathcal{L}_{LCS} + \lambda_2 \mathcal{L}_{SC},
\end{equation}
where $\lambda_1$ and $\lambda_2$ are hyperparameters balancing regularization against the primary task objective. Specifically, enforcing sparsity constraints during the selection of higher-order neighbors per hop allows the model to strike a balance between preserving critical features and mitigating redundant information.

Overall, this optimization framework ensures that our \model maintains a balance between node classification performance, robust high-order neighbor aggregation, and computational efficiency. The algorithm of our ScaleGNN and time complexity analysis are in the Appendix sections \ref{algorithm_appendix} and \ref{time_appendix}, respectively.

%% file: Tables/Datasets.tex
\begin{table*}[t]
\centering
\setlength{\tabcolsep}{9pt}  
\renewcommand{\arraystretch}{1}  
\caption{Statistical summaries of datasets.}
\vspace{-2mm}
\label{table1}
\begin{tabular}{c|c|c|c|c|c|c|c}
\toprule
\multicolumn{2}{c|}{Datasets} & \#Nodes & \#Edges & \#Features & \#Classes & \#Train/Val/Test & Description\\
\midrule
\multicolumn{2}{c|}{Citeseer} & 3,327 & 4,732 & 3,703 
& 6 & 120/500/1000 & citation network\\ 
\multicolumn{2}{c|}{Cora} & 2,708 & 5,429 & 1,433 & 7
& 140/500/1000 & citation network\\ 
\multicolumn{2}{c|}{Pubmed} & 19,717 & 44,338 & 500 & 3
& 60/500/1000 & citation network \\
\midrule
\multicolumn{2}{c|}{ogbn-arxiv} & 169,343 & 1,166,243 & 128 & 40 & 91k/30k/49k & citation network\\ 
\multicolumn{2}{c|}{ogbn-products} & 2,449,029 & 61,859,140 & 100 & 47 & 196k/49k/2,204k & co-purchasing network\\ 
\multicolumn{2}{c|}{ogbn-papers100M} & 111,059,956 & 1,615,685,872 & 128 & 172 & 1200k/200k/146k & citation network\\ 

\bottomrule
\end{tabular}
\label{table_data}
\vspace{-4mm}
\end{table*}

%% file: Main/5_Experiment.tex
\section{Experiments}
In this section, we evaluate the performance of our proposed method through extensive experiments and answer the following questions: 
\begin{itemize}[leftmargin=*]
    \item \textbf{(RQ1)} Can \model outperform SOTA GNN methods regarding predictive accuracy on real-world datasets?
    \item \textbf{(RQ2)} How does the efficiency of \model compare to other baseline methods? 
    \item \textbf{(RQ3)} How dose \model mitigate the over-smoothing issue?
    \item \textbf{(RQ4)} What are the effects of different modules in \model? 
    \item \textbf{(RQ5)} How do different hyperparameter settings affect ScaleGNN? 
    \item \textbf{(RQ6)} How does \model achieve the trade-off between efficiency and accuracy? 
\end{itemize}
\subsection{Experimental Settings}

\subsubsection{Datasets}
We have evaluated the effectiveness of ScaleGNN using six real-world graph datasets, three small-scale datasets Citeseer, Cora, and Pubmed, and three large-scale datasets ogbn-arxiv, ogbn-products, and ogbn-papers100M. The detailed introduction of the used datasets is in the Appendix section~\ref{dataset_appendix}.

\subsubsection{Baselines}
We compare our method with three categories of GNN baselines as follows: (1) traditional GNNs: R-GCN~\cite{schlichtkrull2018modeling}, Cluster-GCN~\cite{chiang2019cluster}, HetGNN~\cite{zhang2019heterogeneous}, HAN~\cite{wang2019heterogeneous}, MAGNN~\cite{fu2020magnn}, Simple-HGN~\cite{lv2021we}, and HINormer~\cite{mao2023hinormer}; (2) deep GNNs: SIGN~\cite{rossi2020sign}, S\textsuperscript{2}GC~\cite{zhu2021simple}, GBP~\cite{chen2020scalable}, GAMLP~\cite{zhang2022graph}, GRAND+~\cite{feng2022grand+}, and LazyGNN~\cite{xue2023lazygnn}; (3) scaleable GNNs: NARS~\cite{yu2020scalable}, SeHGNN~\cite{yang2023simple}, RpHGNN~\cite{hu2024efficient}, and TOP~\cite{shiaccurate2025}. The detailed description of all baselines is in the Appendix section~\ref{baseline_appendix}.

\subsubsection{Implementation Details}
The detailed implementation of our \model can be found in the Appendix section~\ref{setting_appendix}.

\input{Tables/Baselines}

\subsection{Performance Comparison (RQ1)}
We evaluate the performance of our \model and all baselines on six real-world datasets under the node classification task. The experimental results are shown in Table~\ref{tab:baselines}. The best results are highlighted in bold, and the second-best results are underlined. As we can see, our \model achieves the optimal performance, significantly outperforming all baselines in both micro-F1 and macro-F1 metrics on six datasets, with specific emphasis on the improvement relative to the SOTA methods. 

Traditional GNN methods work flawlessly on small-scale datasets, but most of them cannot be extended to large-scale graphs and are prone to OOM issues. For example, R-GCN, HAN, Simple-HGN, and HINormer can get experimental results on ogbn-arxiv, but cannot run on larger-scale graphs such as ogbn-products and ogbn-paper100M. For deep GNN methods, they work on learning large-scale graph representations using deeper GNNs while mitigating the over-smoothing problem on large-scale graphs. A common issue with such methods is that they typically convolve dozens or even tens of GNN layers. While capturing the caveat features to some extent, it also greatly increases the computational complexity of the model. Pre-computation-based GNNs are distinguished from end-to-end models by the fact that they consume a significant amount of time in pre-computing the tensor of large-scale graphs, so that the models do not need to iteratively learn graph representations during training and inference.

It is worth mentioning that our basic model ScaleGNN\textsubscript{b}, although retaining only the basic adaptive high-order feature fusion module, achieves a very similar performance to SOTA methods and runs efficiently, far outperforming all baselines. ScaleGNN gains further performance despite being less efficient than ScaleGNN\textsubscript{b}. Therefore, ScaleGNN and ScaleGNN\textsubscript{b} provide an important reference value for us to explore the trade-off between model accuracy and efficiency. 

\subsection{Efficiency Analysis (RQ2)}

\begin{figure}[t]
    \centering
    \vspace{-2mm}
    \includegraphics[width=0.49\textwidth]{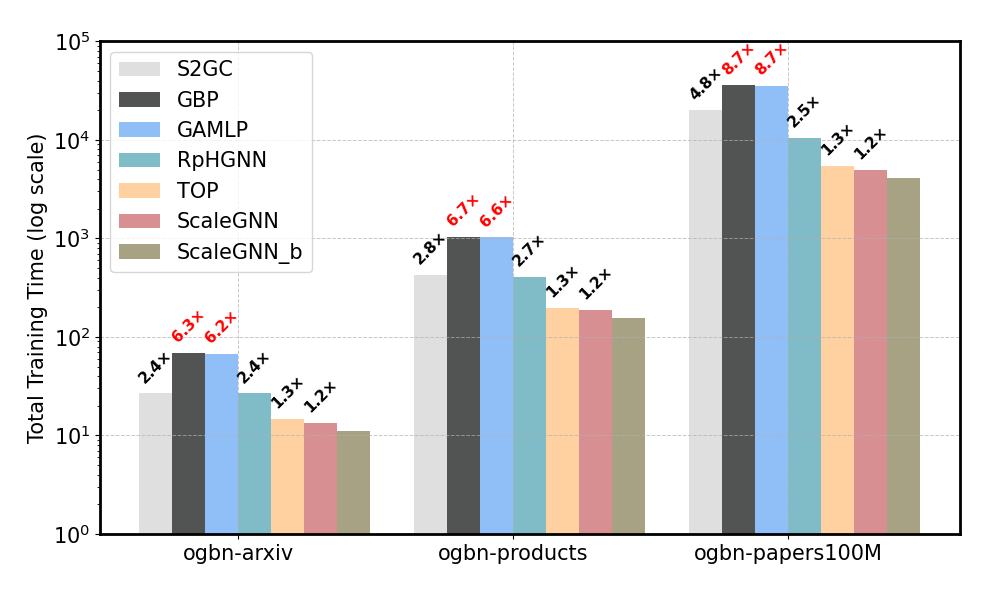}
    \vspace{-8mm}
	\caption{Efficiency comparison with SOTA GNN methods, and speedup analysis compared with our ScaleGNN\textsubscript{b}.
    }
	\label{fig:efficiency}
    \vspace{-6mm}
\end{figure}

To evaluate the efficiency of our proposed methods, ScaleGNN and ScaleGNN\textsubscript{b}, we compare their total runtime with several state-of-the-art baselines, including S\textsuperscript{2}GC, GBP, GAMLP, and RpHGNN, across three benchmark datasets: ogbn-arxiv, ogbn-products, and ogbn-papers100M. The results presented in Fig.~\ref{fig:efficiency} clearly demonstrate the superior efficiency of ScaleGNN and ScaleGNN\textsubscript{b} over all competing methods.

\subsubsection{Computation Efficiency}
As shown in the figure, ScaleGNN\textsubscript{b} achieves the lowest runtime across all datasets, followed closely by ScaleGNN. Specifically, on the large-scale ogbn-papers100M dataset, ScaleGNN\textsubscript{b} completes the model running in only 4,110.2 seconds, significantly outperforming S\textsuperscript{2}GC (19,914.4s), GBP (35,706.1s), GAMLP (35,656.6s), and RpHGNN (10,399.8s). This trend is also consistently observed across ogbn-arxiv and ogbn-products datasets, where ScaleGNN\textsubscript{b} reduces runtime by at least 4.1× to 6.7× compared to the best-performing baseline. The significant reduction in runtime can be attributed to the efficient design of ScaleGNN, which optimizes neighborhood aggregation and selectively samples informative neighbors, thereby reducing computational overhead. Moreover, the enhanced scalability of ScaleGNN\textsubscript{b} further refines this efficiency by introducing additional optimizations in feature propagation and memory management.

\subsubsection{Contrastive Speedup Analysis}
To better illustrate the efficiency advantage, we annotate each baseline method with its relative speedup factor compared to ScaleGNN\textsubscript{b}. It is evident that methods like GBP and GAMLP exhibit substantially higher computational costs, with runtime exceeding 6× that of ScaleGNN\textsubscript{b} in certain cases. Even RpHGNN, which demonstrates competitive efficiency, remains 2.5× to 3.1× slower than ScaleGNN\textsubscript{b} on large datasets. These results highlight the capability of ScaleGNN to effectively mitigate the scalability limitations of existing methods. By reducing redundant computations and dynamically adjusting the information flow, our approach enables substantial efficiency improvements without compromising accuracy. The consistent speedup across diverse datasets further underscores the robustness of our design, making it particularly suitable for large-scale graph representation learning tasks.

In summary, ScaleGNN and ScaleGNN\textsubscript{b} achieve remarkable computational efficiency, substantially reducing runtime compared to existing approaches. The combination of selective neighbor aggregation, adaptive feature diffusion, and memory-efficient propagation mechanisms enables our methods to handle large-scale graphs with minimal computational overhead. These findings underscore the practical applicability of ScaleGNN in real-world scenarios where the model efficiency is a critical concern.

\subsection{Mitigating Over-smoothing Issue (RQ3)}
Over-smoothing is a well-known issue in deep graph neural networks, where node representations become indistinguishable as the number of propagation layers increases. This problem is particularly severe in large-scale graphs such as ogbn-papers100M, where excessive feature mixing leads to performance degradation. To evaluate the effectiveness of our approach in addressing this challenge, we compare ScaleGNN with several SOTA baselines, including S\textsuperscript{2}GC, GBP, GAMLP, and RpHGNN, across different propagation depths.
\begin{figure}[h]
    \centering
    \captionsetup[subfigure]{font=small}  
    \subfloat[ogbn-products]{
    \label{fig:o1}
    \includegraphics[scale=0.2]{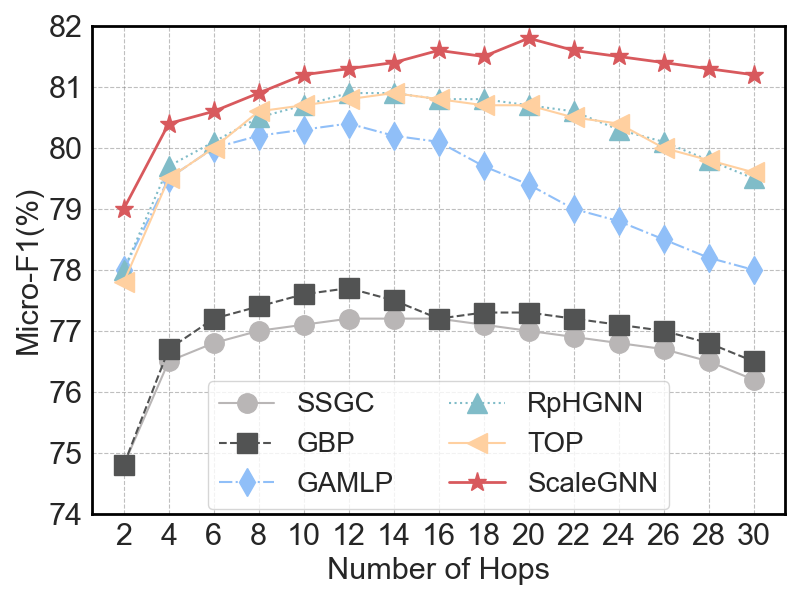}
    }
    \hspace{-3mm}
    \subfloat[ogbn-paper100M]{
    \label{fig:o2}
    \includegraphics[scale=0.2]{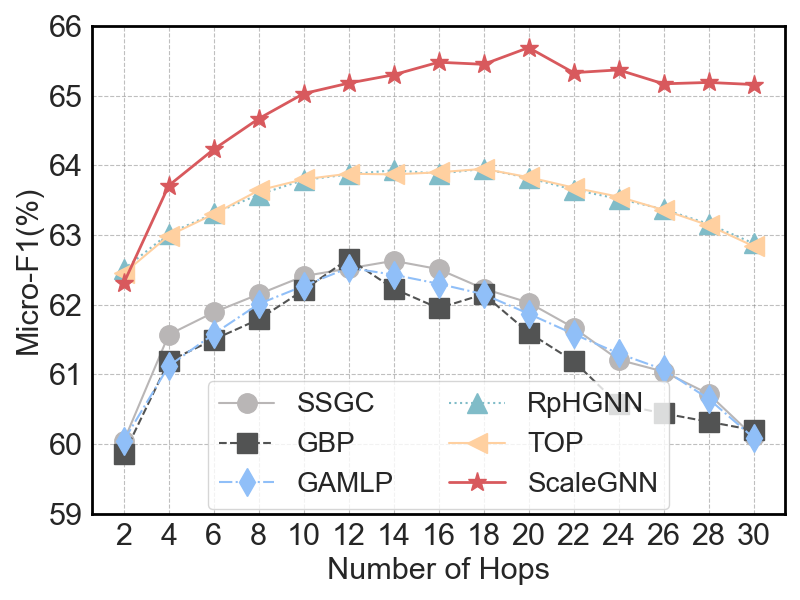}
    }
    \vspace{-2mm}
    \caption{Experimental results of our model over SOTA GNN models \textit{w.r.t.} the number of hops.
    }
    \label{fig:oversmoothing}
    \vspace{-4mm}
\end{figure}

Fig.~\ref{fig:oversmoothing} shows that traditional GNNs experience significant performance drops as the number of propagation layers increases. For example, S\textsuperscript{2}GC, GBP, and GAMLP achieve their peak performance at moderate depths (around 8-12 layers) but degrade rapidly beyond this point. This phenomenon occurs because deeper layers cause excessive aggregation, leading to the loss of discriminative node features. In contrast, ScaleGNN exhibits a more stable performance curve. Even at greater depths (\eg, 30 layers), ScaleGNN maintains a high micro-F1 score, outperforming all baselines, which demonstrates that \model effectively preserves structural and feature information, preventing the collapse of node representations.

The key to ScaleGNN’s resilience against over-smoothing lies in its adaptive feature aggregation mechanism. Unlike conventional approaches that uniformly aggregate neighborhood information, ScaleGNN dynamically balances local and global features, ensuring that useful information is retained while mitigating excessive smoothing. Specifically, our \model selectively incorporates higher-order neighbors with controlled influence, preventing over-mixing of features. It employs adaptive diffusion mechanisms to regulate information propagation, ensuring a balance between short-term and long-term dependencies. \model utilizes LCS to prioritize informative neighbors, reducing noise and redundancy in feature aggregation. These strategies allow ScaleGNN to maintain a richer set of node representations, even in deep architectures, making it particularly effective for large-scale graphs.

\subsection{Ablation Study (RQ4)}
To evaluate the effectiveness of each component in \model, we further conduct ablation studies on different variants. Specifically, we generate three variants as follows:
\begin{itemize}[leftmargin=*]
    \item \textbf{\textit{w/o Ada}} removes the adaptive high-order feature fusion module and employs the deep GNN method S\textsuperscript{2}GC~\cite{zhu2021simple} to obtain the graph representation.
    \item \textbf{\textit{w/o Low}} excludes the low-order enhanced feature aggregation and only employs adaptive high-order feature fusion.
    \item \textbf{\textit{w/o LCS}} removes the high-order redundant feature masking.
    \item \textbf{\textit{w/o SC}} removes the masking sparse constraint and keeps the same retained number of nodes in each $\mathbf{A}_k^{\text{filter}}$.
\end{itemize}
\vspace{-4mm}
\input{Tables/Ablation_large}
The results demonstrate that each component is vital for improving ScaleGNN’s performance. \textbf{\textit{w/o Ada}} variant experiences the largest performance drop across all datasets, with reductions of 3.07\% in Micro-F1 and 2.17\% in Macro-F1 on ogbn-arxiv, 2.07\% and 1.14\% on ogbn-products, and 1.24\% and 1.44\% on ogbn-paper100M, highlighting the importance of adaptive high-order feature fusion for capturing structural dependencies and mitigating information loss. \textbf{\textit{w/o Low}} variant also shows notable degradation, especially on ogbn-arxiv and ogbn-paper100M, with Micro-F1 dropping by 1.33\% and 0.72\%, respectively, indicating that low-order enhanced feature aggregation is crucial for maintaining local information. \textbf{\textit{w/o LCS}} variant sees moderate degradation, with Micro-F1 decreasing over three large-scale datasets, showing the effectiveness of high-order redundant feature masking in reducing redundancy and preventing excessive feature mixing. \textbf{\textit{w/o SC}} shows that it still falls short of the full-fledged ScaleGNN model despite being the best of all variants, demonstrating the important role of the LCS-based sparse constraints. It employs an attention-based relevance function, which allows \model to adaptively assess and filter neighbors in a differentiable and task-driven manner, significantly improving its generalization and scalability in large-scale graphs.

Overall, \model consistently outperforms all variants, confirming the significant contribution of each component. Adaptive high-order feature fusion is the most influential in improving performance, followed by low-order enhanced feature aggregation and high-order redundant feature masking. The combination of these mechanisms enables ScaleGNN to achieve superior results on various large-scale graph datasets. 

\begin{figure}[t]
	\centering 
    \captionsetup[subfigure]{font=small}  
    \subfloat[ogbn-products]{
    \label{fig:b1}
    \includegraphics[width=4cm, height=1.9cm]{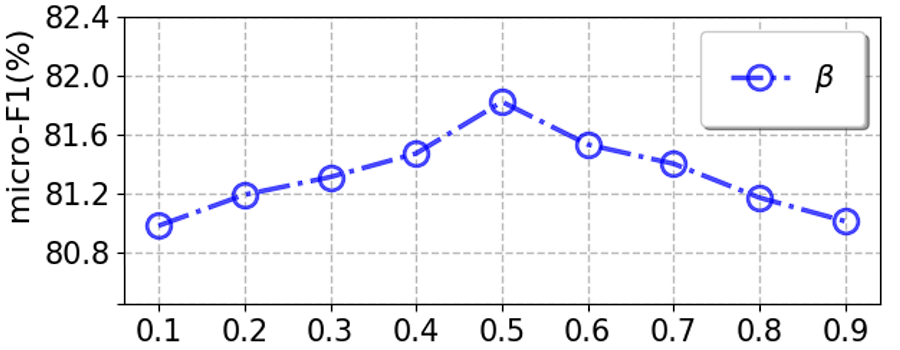}
    }
    \hspace{-3mm}
    \subfloat[ogbn-papers100M]{
    \label{fig:b2}
    \includegraphics[width=4cm, height=1.9cm]{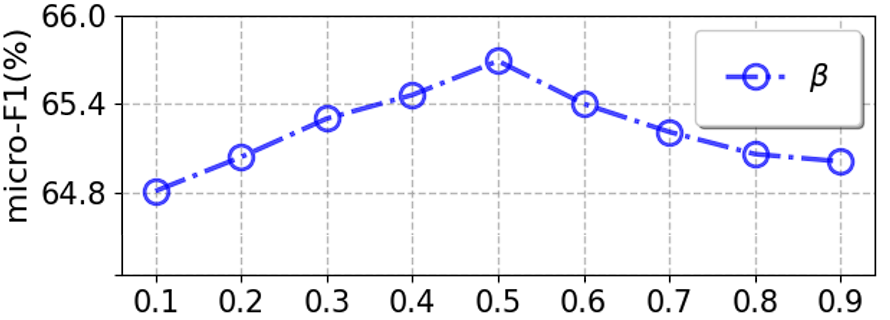}
    }
    \vspace{-2mm}
    \caption{The effect of balanced hyperparameter $\beta$.}
    \label{fig:parameter}
    \vspace{-4mm}
\end{figure}
\begin{figure}[t]
	\centering 
    \captionsetup[subfigure]{font=small}  
    \subfloat[ogbn-products]{
    \label{fig:l1}
    \includegraphics[width=4cm, height=1.9cm]{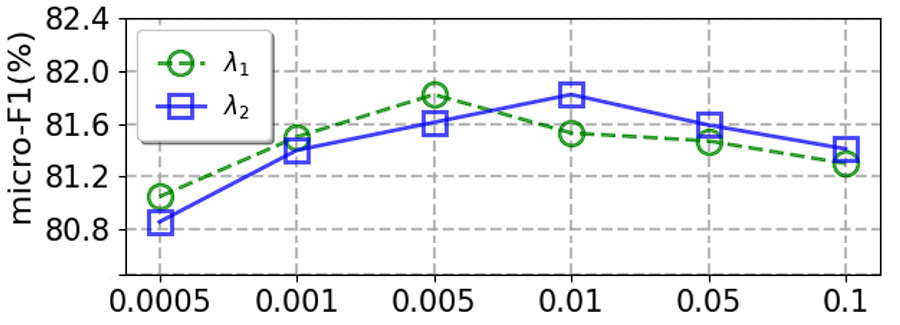}
    }
    \hspace{-3mm}
    \subfloat[ogbn-papers100M]{
    \label{fig:l2}
    \includegraphics[width=4cm, height=1.9cm]{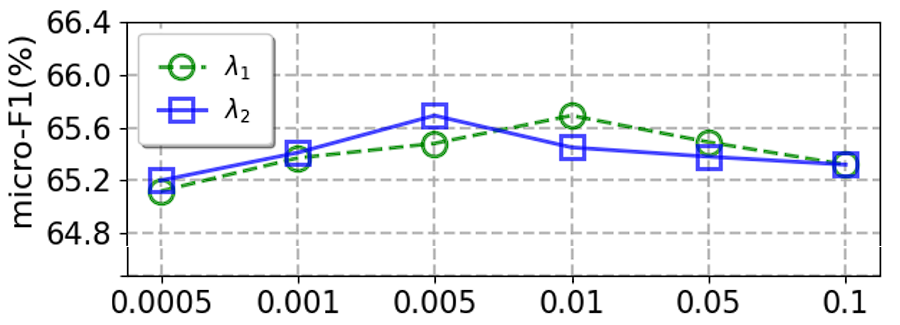}
    }
    \vspace{-2mm}
    \caption{The effect of loss fuction weights $\lambda_1$ and $\lambda_2$.}
    \label{fig:parameter}
    \vspace{-6mm}
\end{figure}
\subsection{Parameter Sensitivity Analysis (RQ5)}
\subsubsection{The effect of hyperparameters $\beta$}
We conduct a sensitivity analysis of $\beta$, which balances low-order and high-order feature aggregations, by varying its value from 0.1 to 0.9 and evaluating the Micro-F1 score (\%). The results in Fig.~\ref{fig:parameter} show that performance improves as $\beta$ increases, peaking at $\beta = 0.5$. Beyond this point, further increases in $\beta$ lead to a gradual decline, likely due to overemphasizing high-order information, which may introduce noise or redundancy. These findings highlight the importance of $\beta$ to strike an optimal balance between local and global information aggregation.

\subsubsection{The effect of hyperparameters $\lambda_1$ and $\lambda_2$}
We assess the sensitivity of $\lambda_1$ and $\lambda_2$ in the loss function by varying their values within \{0.0005, 0.001, 0.005, 0.01, 0.05, 0.1\} and evaluating Micro-F1. As shown in Fig.~\ref{fig:parameter}, \model achieves the best performance at $\lambda_1 = 0.005$, with results improving up to this point and declining thereafter, reflecting the benefit of moderate regularization. In contrast, performance with respect to $\lambda_2$ is more stable, peaking at $\lambda_2 = 0.01$ with only slight variations, indicating its role in maintaining model stability. These results highlight the need for careful tuning: $\lambda_1$ requires precise adjustment for optimal results, while $\lambda_2$ ensures robustness. The loss terms weighted by $\lambda_1$ and $\lambda_2$ complement each other, balancing performance and efficiency.

\subsection{Trade-off Analysis: Performance vs. Efficiency (RQ6)}
In this section, we evaluate the trade-off between efficiency and accuracy for ScaleGNN and ScaleGNN\textsubscript{b} by analyzing their runtime and Micro-F1 scores across diverse hop settings.

\begin{figure}[h]
    \centering
    \captionsetup[subfigure]{font=small}  
    \subfloat[ogbn-products]{
    \label{fig:o3}
    \includegraphics[scale=0.2]{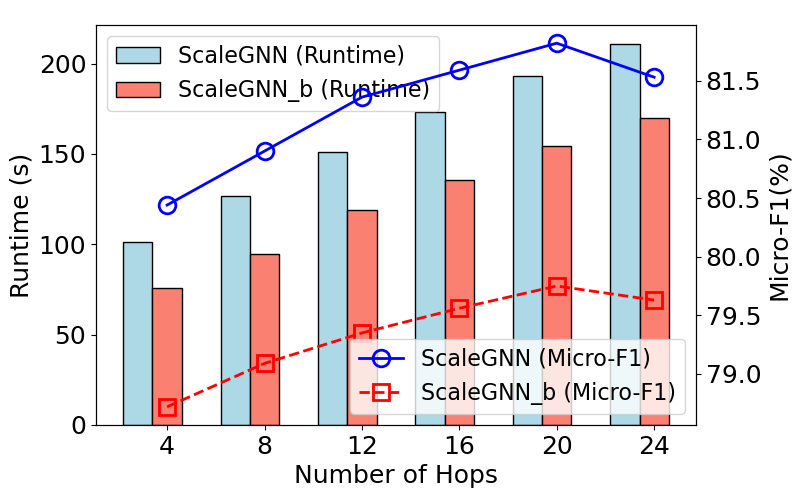}
    }
    \hspace{-3mm}
    \subfloat[ogbn-papers100M]{
    \label{fig:o4}
    \includegraphics[scale=0.2]{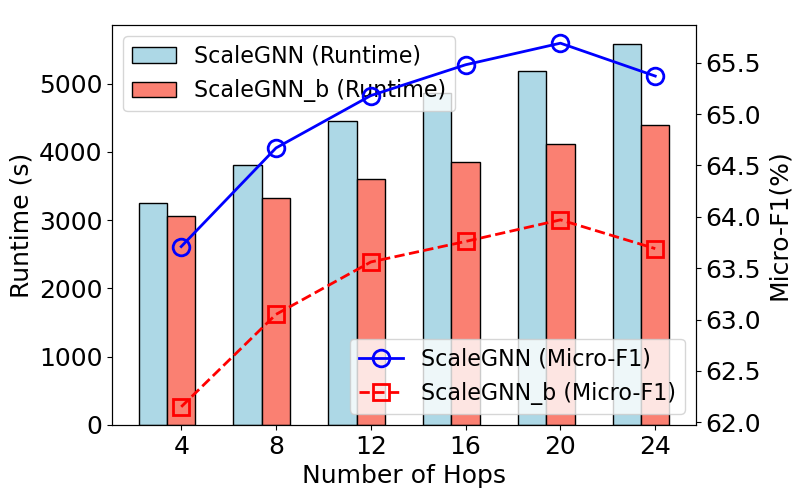}
    }
    \vspace{-2mm}
    \caption{The trade-off between efficiency and accuracy on large-scale graph datasets.}
    \label{fig:tradeoff}
    \vspace{-2mm}
\end{figure}
The runtime analysis reveals in Fig.~\ref{fig:tradeoff} that ScaleGNN incurs a higher computational cost than ScaleGNN\textsubscript{b} across all hop settings. As the number of hops increases, both methods show rising runtimes, reflecting the increased complexity of message passing over larger neighborhoods. However, ScaleGNN consistently requires more computational resources, suggesting that its modeling choices introduce additional overhead. ScaleGNN consistently outperforms ScaleGNN\textsubscript{b} in Micro-F1 scores. Both methods benefit from increasing hops, as it helps capture structural and semantic relationships. However, beyond a certain point, performance gains diminish due to redundant or noisy information from distant nodes. These results highlight the trade-off between accuracy and efficiency. While ScaleGNN achieves higher Micro-F1 scores, it comes with a significantly higher computational cost. In contrast, ScaleGNN\textsubscript{b} offers a more efficient alternative with a slight performance compromise. This suggests that model selection should be based on practical constraints: ScaleGNN is ideal for accuracy-focused tasks, while ScaleGNN\textsubscript{b} is better for resource-constrained environments. 

%% file: Tables/Baselines.tex
\begin{table*}[t]
\centering
\caption{Performance comparison of all models on six datasets. 
Mi-F1 and Ma-F1 are short for Micro-F1 and Macro-F1. Marker * indicates the results are statistically significant against the best-performed baselines (t-test with p-value $<$ 0.01).
}
\vspace{-2mm}
\setlength{\tabcolsep}{1.3mm}{}	
\begin{tabular}{cc|cc|cc|cc|cc|cc|cc}
\toprule
\multicolumn{2}{c|}{\multirow{2}{*}{Method}}  & \multicolumn{2}{c|}{Citeseer} & \multicolumn{2}{c|}{Cora}  & \multicolumn{2}{c|}{Pubmed}  & \multicolumn{2}{c|}{ogbn-arxiv} & \multicolumn{2}{c|}{ogbn-products} & \multicolumn{2}{c}{ogbn-paper100M}\\
 &  & Mi-F1 & Ma-F1 & Mi-F1 & Ma-F1 & Mi-F1 & Ma-F1 & Mi-F1 & Ma-F1 & Mi-F1 & Ma-F1 & Mi-F1 & Ma-F1 \\
\midrule
\multirow{6}{*}{Trad. GNN} & RGCN & 70.53 & 70.14 & 81.23 & 80.59 & 78.37 & 75.58 & 58.26 & 46.83 & 48.94 & 48.04 & 42.31 & 31.12 \\
& Cluster-GCN & 71.25 & 70.72 & 82.10 & 80.88 & 78.84 & 76.06 & 78.96 & 75.95 & 71.59 & 63.88 & 62.34 & 54.02 \\
& HetGNN & 70.55 & 70.13 & 81.20 & 80.57 & 78.40 & 75.66 & OOM. & OOM. & OOM. & OOM. & OOM. & OOM. \\
& HAN & 71.10 & 70.57 & 81.39 & 80.86 & 78.53 & 75.85 & 54.68 & 31.50 & OOM. & OOM. & OOM. & OOM.  \\
& MAGNN & 71.22 & 70.69 & 81.57 & 80.98 & 78.55 & 76.01 & OOM. & OOM. & OOM. & OOM. & OOM. & OOM.  \\
& Simple-HGN & 71.34 & 70.78 & 81.88 & 81.14 & 78.88 & 76.17 & 69.47 & 58.29 & OOM. & OOM. & OOM. & OOM.  \\
& HINormer & 72.75 & 71.91 & 82.19 & 81.67 & 79.35 & 76.60 & 70.71 & 59.68 & OOM. & OOM. & OOM. & OOM. \\
\midrule
\multirow{6}{*}{Deep GNN} & SIGN & 72.44 & 71.97 & 82.11 & 80.46 & 79.53 & 76.44 & 71.78 & 62.99 & 78.21 & 70.10 & 62.32 & 53.59 \\
& S\textsuperscript{2}GC & 72.78 & 72.13 & 82.45 & 82.09 & 79.56 & 76.67 & 71.88 & 63.23 & 77.09 & 69.54 & 62.63 & 53.80 \\
& GBP & 72.61 & 72.08 & 83.42 & \underline{82.42} & 80.56 & 79.13 & 71.42 & 62.41 & 77.27 & 69.49 & 62.66 & 53.97 \\
& GAMLP & 72.51 & 72.06 & 82.31 & 80.32 & 79.11 & 76.15 & 71.88 & 63.11 & 80.34 & 71.34 & 62.53 & 53.79 \\
& GRAND+ & 72.48 & 72.03 & 82.04 & 80.23 & 79.24 & 76.46 & 71.43 & 62.30 & 77.52 & 69.70 & 61.85 & 53.49 \\
& LazyGNN & 72.91 & 72.21 & 82.45 & 80.87 & 79.92 & 77.03 & 71.63 & 62.72 & 77.60 & 69.90 & 61.97 & 53.55 \\
\midrule
\multirow{3}{*}{Scal. GNN} & NARS & 72.41 & 71.21 & 81.79 & 80.30 & 78.82 & 75.82 & 70.75 & 59.88 & 78.08 & 70.20 & 62.52 & 53.63 \\
& SeHGNN & 72.90 & 72.45 & 83.04 & 81.43 & 79.71 & 77.70 & 71.82 & 62.69 & 79.68 & 70.85 & 63.75 & 54.74 \\
& RpHGNN & \underline{73.12} & 72.54 & 83.27 & 82.37 & 80.58 & 79.32 & \underline{72.06} & \underline{63.34} & 80.89 & 71.44 & 63.95 & 55.50 \\
& TOP & 73.11 & \underline{72.69} & 83.30 & 82.35 & \underline{80.60} & \underline{79.35} & 71.82 & 62.90 & \underline{80.93} & \underline{71.45} & 63.95 & 55.48 \\
\midrule
\multirow{2}{*}{Ours} & ScaleGNN\textsubscript{b} & 72.95 & 72.47 & \underline{83.42} & 82.40 & 80.48 & 79.15 & 71.68 & 62.65 & 79.75 & 70.87 & \underline{63.97} & \underline{55.60} \\
& \textbf{ScaleGNN} & \textbf{74.44*} & \textbf{73.76*} & \textbf{84.75*} & \textbf{84.21*} & \textbf{82.11*} & \textbf{80.67*} & \textbf{73.25*} & \textbf{64.20*} & \textbf{81.82*} & \textbf{72.47*} & \textbf{65.69*} & \textbf{57.42*}\\
\bottomrule
\end{tabular}
\vspace{-2mm}
\label{tab:baselines}
\end{table*}

%% file: Tables/Ablation_large.tex
\begin{table}[h]
\centering
\caption{The comparison of ScaleGNN and its variants in terms of Mi-F1 and Ma-F1 on three large-scale datasets. }
\vspace{-2mm}
\label{tab:ablation}
\setlength{\tabcolsep}{1.4mm}{}	
\begin{tabular}{c|cc|cc|cc} 
\toprule
 Dataset & \multicolumn{2}{c|}{ogbn-arxiv} & \multicolumn{2}{c|}{ogbn-products} & \multicolumn{2}{c}{ogbn-paper100M}  \\
 Metrics & Mi-F1 & Ma-F1 & Mi-F1 & Ma-F1 & Mi-F1 & Ma-F1 \\
\midrule
\midrule
\textbf{\textit{w/o Ada}} & 70.13 & 62.03 & 79.82 & 70.31 & 64.19 & 55.87  \\
\textbf{\textit{w/o Low}} & 71.92 & 62.85 & 79.97 & 71.22 & 64.97 & 56.56  \\
\textbf{\textit{w/o LCS}} & 71.74 & 62.17 & 80.08 & 70.83 & 64.45 & 55.98  \\
\textbf{\textit{w/o SC}} & 72.52 & 63.87 & 81.26 & 71.77 & 65.46 & 57.19  \\
\midrule
\textbf{ScaleGNN} & \textbf{73.25} & \textbf{64.20} & \textbf{81.82} & \textbf{72.47} & \textbf{65.69} & \textbf{57.42} \\
\bottomrule
\end{tabular}
\end{table}

%% file: Main/6_Conclusion.tex
\section{Conclusion}
We propose a novel scalable GNN model named ScaleGNN, tackling over-smoothing and scalability challenges in large-scale graphs. ScaleGNN adaptively aggregates informative high-order neighbors while suppressing redundancy through a Local Contribution Score (LCS)-based masking mechanism. Extensive experiments show that ScaleGNN outperforms state-of-the-art deep and scalable GNNs in both accuracy and efficiency. Future work includes extending it to heterogeneous graphs and optimizing large-scale training.

%% file: Main/7_Appendix.tex
\appendix
\section{Supplement}


\subsection{The Learning Process of \model}\label{algorithm_appendix}
Algorithm~\ref{al:ScaleGNN} shows the pseudo-code of \model framework. The algorithm summarizes the overall process of the proposed ScaleGNN framework. The model first constructs pure $k$-hop adjacency matrices to isolate neighbor information at different topological scales. It then performs adaptive high-order feature fusion by learning attention weights across multiple hops, and separately aggregates low-order features via the original adjacency. To suppress redundant information introduced by distant neighbors, a local contribution score (LCS) is used to select the most informative nodes, refining the high-order structure. Finally, the node representations are obtained by weighted fusion of the low- and high-order embeddings.
\input{Tables/Algorithm}


\subsection{Time Complexity Analysis}\label{time_appendix}
To evaluate the computational efficiency of \model, we analyze its time complexity and compare it with existing scalable GNNs. The overall complexity consists of three main components: pre-processing, training, and inference. 

Let $|\mathcal{V}|$ be the number of nodes, $|\mathcal{E}|$ be the number of edges, and $f$ be the feature dimension, $K$ be the maximum number of hops, $n$ be the maximum number of top-$m_i$ ($i \in \{1,2,\ldots,K\}$) neighbors. $L$ is the number of layers in MLP classifiers of SIGN, GAMLP, and GBP, $S$ in NARS denotes the number of subgraphs, $P$ in SeHGNN denotes the number of divided groups, and $R$ in RpHGNN is the number of node classes. In our proposed \model, we perform the following major computations:

\begin{itemize}[leftmargin=*]
    \item \textbf{Multi-hop Adjacency Construction}: To construct order-distilled adjacency matrices $\{\mathbf{A}_i\}_{i=1}^{K}$ where $\mathbf{A}_i = \mathbf{A}^i - \mathbf{A}^{i-1}$, we perform $K$ sparse matrix multiplications. The cost per multiplication is $O(|\mathcal{E}|)$ in sparse setting, so the total cost is $O(K|\mathcal{E}|)$.
    
    \item \textbf{LCS-based Neighbor Filtering}: 
    For each node $v$ and $k$-hop neighbor $u$, attention-based LCS involves two linear projections ($\mathbf{W}_1\mathbf{x}_v$, $\mathbf{W}_2\mathbf{x}_u$), a dot-product, and softmax computation: (1) Projection cost: amortized $O(|\mathcal{V}|fd_f)$ (precomputed). (2) Attention score computation: $O(K|\mathcal{V}|nd_f)$. (3) The top-$m_k$ filtering: $O(K|\mathcal{V}|n\log n)$ using efficient heap sort. The overall complexity is approximately $O(Kn|\mathcal{V}|f)$ and remains linear in retained neighbors and supports batch-wise parallel execution.

    \item \textbf{Feature Aggregation and Training}: 
    Once filtered adjacency matrices are obtained, the time complexity of aggregation step is $O(Kn|\mathcal{V}|f)$. The time complexity of training and inference is $O(|\mathcal{V}|f^2)$.
\end{itemize}

Finally the end-to-end time complexity of our \model is approximately $O(K|\mathcal{E}| + Kn|\mathcal{V}|f + |\mathcal{V}|f^2$). Table~\ref{tab:time_complexity} compares the time complexity of ScaleGNN with other scalable GNN architectures.

\begin{table}[h]
    \centering
    \caption{Time complexity comparison of existing deep and scalable GNN models.}
    \label{tab:time_complexity}
    \setlength{\tabcolsep}{0.8mm}{}
    \begin{tabular}{lccc}
        \toprule
        \textbf{Method} & \textbf{Pre-processing} & \textbf{Training} & \textbf{Inference} \\
        \midrule
        S\textsuperscript{2}GC & \( \mathcal{O}(K|\mathcal{E}|f) \) & \( \mathcal{O}(|\mathcal{V}|f^2) \) & \( \mathcal{O}(|\mathcal{V}|f^2) \) \\
        SIGN & \( \mathcal{O}(K|\mathcal{E}|f) \) & \( \mathcal{O}(L|\mathcal{V}|f^2) \) & \( \mathcal{O}(L|\mathcal{V}|f^2) \) \\
        GAMLP & \( \mathcal{O}(K|\mathcal{E}|f) \) & \( \mathcal{O}(L|\mathcal{V}|f^2) \) & \( \mathcal{O}(L|\mathcal{V}|f^2) \) \\
        GBP & \( \mathcal{O}(K|\mathcal{E}|f + K \sqrt{|\mathcal{E}|} \log|\mathcal{V}|) \) & \( \mathcal{O}(L|\mathcal{V}|f^2) \) & \( \mathcal{O}(L|\mathcal{V}|f^2) \) \\
        \midrule
        NARS & $\mathcal{O}(K|\mathcal{V}|Sf+K|\mathcal{V}|fd_f)$ & $\mathcal{O}(|\mathcal{V}|f^2)$ & $\mathcal{O}(|\mathcal{V}|f^2)$ \\
        SeHGNN & $\mathcal{O}(K|\mathcal{V}|Pf+P^2|\mathcal{V}|fd_f)$ & $\mathcal{O}(|\mathcal{V}|f^2)$ & $\mathcal{O}(|\mathcal{V}|f^2)$ \\
        RpHGNN & $\mathcal{O}(K|\mathcal{V}|Rf+R|\mathcal{V}|fd_f)$ & $\mathcal{O}(|\mathcal{V}|f^2)$ & $\mathcal{O}(|\mathcal{V}|f^2)$ \\
        \textbf{ScaleGNN} & \( \mathcal{O}(Kn|\mathcal{V}|f) + K|\mathcal{E}|)\) & \( \mathcal{O}(|\mathcal{V}|f^2) \) & \( \mathcal{O}(|\mathcal{V}|f^2) \) \\
        \bottomrule
    \end{tabular}
    \vspace{-2mm}
\end{table}

Compared to traditional GNNs that suffer from excessive computation due to repeated feature propagation, ScaleGNN optimizes computational efficiency in two ways: (1) \textbf{Sparse high-hop neighbor selection}: By leveraging LCS-based filtering, ScaleGNN significantly reduces the number of unnecessary high-hop neighbors involved in feature propagation, improving computational efficiency. (2) \textbf{Lightweight aggregation}: Since ScaleGNN explicitly separates low-hop and high-hop features, it avoids redundant computations, leading to lower memory and runtime costs.

Empirical results on real-world datasets demonstrate that the proposed ScaleGNN achieves faster convergence and lower training time compared to existing model-simplification methods while maintaining high predictive accuracy.

\subsection{Datasets}\label{dataset_appendix}
The first three datasets (Citeseer, Cora, and Pubmed) are relatively small and have been adopted by existing work~\cite{liang2025towards}, while the latter three (ogbn-arxiv, ogbn-products, and ogbn-papers100M) are large-scale heterogeneous graphs commonly used in scalable HGNN evaluations. The detailed dataset description is shown in Table~\ref{table_data}.

\subsection{Baselines}\label{baseline_appendix}
We compare our MHCL with the following 12 baselines, which can be deeply divided into four categories:

\noindent
\textbf{\textit{Traditional GNN methods:}}
\begin{itemize}[leftmargin=*]
\item \textbf{R-GCN~\cite{schlichtkrull2018modeling}} introduces relation-specific graph convolution operations to effectively model multi-relational knowledge graphs and achieves strong performance on tasks like link prediction and entity classification.
\item \textbf{Cluster-GCN~\cite{chiang2019cluster}} is an efficient GCN algorithm that trains models by sampling dense subgraphs partitioned via graph clustering. It significantly reduces memory and computational overhead, enabling efficient training on million-scale graphs for the first time.
\item \textbf{HetGNN~\cite{zhang2019heterogeneous}} is a heterogeneous graph neural network that jointly learns from both heterogeneous graph structure and node content by aggregating multi-type neighbor and attribute information, achieving superior performance on tasks like link prediction and node classification.
\item \textbf{HAN~\cite{wang2019heterogeneous}} employs a hierarchical attention mechanism, combining node-level and semantic-level attention to effectively capture the importance of both meta-path neighbors and meta-paths, enabling expressive node representations in heterogeneous graphs. 
\item \textbf{MAGNN~\cite{fu2020magnn}} integrates node content, intermediate nodes along meta-paths, and multiple meta-paths through hierarchical aggregation, enabling more expressive and accurate representations in heterogeneous graphs.
\item \textbf{Simple-HGN~\cite{lv2021we}} is a strong and straightforward baseline for heterogeneous graph learning that, with proper input processing, outperforms previous heterogeneous GNN models across standardized benchmark datasets.
\item \textbf{HINormer~\cite{mao2023hinormer}} leverages a graph transformer architecture with global-range attention and specialized encoders to effectively capture both structural and heterogeneous relational information, enabling comprehensive node representations in heterogeneous information networks.
\end{itemize}

\noindent
\textbf{\textit{Deep GNN methods:}}
\begin{itemize}[leftmargin=*]
    \item \textbf{SIGN~\cite{rossi2020sign}} eliminates the need for neighbor sampling by using efficiently precomputed graph convolutional filters of varying sizes, enabling fast and scalable training and inference on large-scale graphs.
    \item \textbf{S\textsuperscript{2}GC~\cite{zhu2021simple}} uses a modified Markov diffusion kernel to efficiently aggregate information from both local and global neighborhoods, capturing multi-scale context while mitigating over-smoothing in deeper graph convolutions.
    \item \textbf{GBP~\cite{chen2020scalable}} employs a localized bidirectional propagation strategy to achieve efficient and scalable graph learning, enabling sub-linear time complexity and state-of-the-art performance on massive graphs without relying on sampling.
    \item \textbf{GAMLP~\cite{zhang2022graph}} adaptively integrates information from multiple neighborhood scales using attention, effectively addressing over-smoothing and enabling highly scalable and efficient graph learning on large-scale and sparse graphs.
    \item \textbf{GRAND+~\cite{feng2022grand+}} enhances scalable graph learning by introducing an efficient pre-computation algorithm for propagation and a confidence-aware consistency loss, enabling fast and accurate semi-supervised learning on large-scale graphs.
    \item \textbf{LazyGNN~\cite{xue2023lazygnn}} captures long-distance dependencies in graphs using shallow models rather than deeper architectures, effectively avoiding neighborhood explosion and achieving efficient, scalable performance on large-scale graphs.
\end{itemize}

\noindent
\textbf{\textit{Scalable GNN methods:}}
\begin{itemize}[leftmargin=*]
    \item \textbf{NARS~\cite{yu2020scalable}} efficiently computes neighbor-averaged features over sampled relation subgraphs, enabling scalable and accurate learning on large heterogeneous graphs without complex end-to-end GNN architectures.
    \item \textbf{SeHGNN~\cite{yang2023simple}} streamlines heterogeneous graph learning by using pre-computed mean aggregation and a single-layer structure with long meta-paths, combined with transformer-based semantic fusion, achieving high accuracy and fast training with low complexity.
    \item \textbf{RpHGNN~\cite{hu2024efficient}} combines efficient pre-computation with low information loss by using random projection squashing and a relation-wise even-odd propagation scheme, enabling fast and accurate learning on heterogeneous graphs.
    \item \textbf{TOP~\cite{shiaccurate2025}} eliminates costly out-of-batch message passing by introducing message invariance to convert it into efficient in-batch propagation, enabling fast and scalable GNN training on large graphs with minimal accuracy loss.
\end{itemize}

Notice that most of the traditional and deep GNN methods encounter out-of-memory (OOM) issues when dealing with large datasets (such as products and paper100M). As for R-GCN, while the original version presents OOM issues with large datasets, it is often used as a baseline for large-scale HGNNs. To adapt R-GCN for large datasets, we follow NARS’ experimental setting and adopt the neighbor sampling strategy used by GraphSAGE~\cite{hamilton2017inductive}. Besides, scalable GNN methods, NARS, SeHGNN, and RpHGNN, are all pre-computation-based methods that differ from other end-to-end models. Most of their model runtime is consumed in obtaining the pre-computation tensor for large-scale graphs, with only a tiny amount of time spent on training and inference.



\subsection{Implementation Details}\label{setting_appendix}
We implement all baseline methods according to their provided code. Generally, the size of the hidden embedding $d_f$ is selected from \{32, 64, 128, 256, 512\}, and the maximum number $K$ of hops is adjustable. For three small datasets, we set the initial epoch number to 100 and select dropout rate, weight decay, and learning rate from \{0.1, 0.2, 0.3, 0.4, 0.5, 0.6\}, \{1e-5, 5e-5, 1e-4, 5e-4, 1e-3, 5e-3, 1e-2\} and \{0.0001, 0.0005, 0.001, 0.005, 0.01\}, respectively. For three large-scale datasets, we set the epoch number to 1000 and the learning rate to 0.001 without weight decay. The embedding dimension $d$ is chosen from \{32, 64, 128, 256, 512\}, and our method uses GCN as the message aggregator. All experiments are conducted on a machine with an AMD EPYC 7402 24-core CPU and NVIDIA GeForce RTX 4090 (24GB Memory) GPUs.

%% file: Tables/Algorithm.tex
\begin{algorithm}[h]
\caption{ScaleGNN}
\label{al:ScaleGNN}
\begin{algorithmic}[1]
\REQUIRE Graph $\mathcal{G}=(\mathcal{V},\mathcal{E},\mathcal{X})$, adjacency $\mathbf{A}$, max order $K$
\ENSURE Node representations $\mathbf{H}$
\STATE \textbf{\textit{Construct pure $i$-hop adjacency:}}
\FOR{$i=1$ to $K$}
    \STATE $\mathbf{A}_{i} = \mathbf{A}^{i} - \mathbf{A}^{i-1}$ 
\ENDFOR
\STATE \textbf{\textit{Adaptive high-order feature fusion:}}
\STATE Learn weights $\alpha_1, \dots, \alpha_K$ via softmax, $\sum_i \alpha_i=1$
\STATE $\tilde{\mathbf{A}} = \sum_{i=1}^K \alpha_i \mathbf{A}_i$
\STATE $\mathbf{H}_{\text{high}} = \sigma(\tilde{\mathbf{A}}\cdot\mathcal{X}\cdot\mathbf{W}_{\text{high}})$
\STATE \textbf{\textit{Low-order feature aggregation:}}
\STATE $\mathbf{H}_{\text{low}} = \sigma(\mathbf{A}^2\cdot\mathcal{X}\cdot \mathbf{W}_{\text{low}})$
\STATE \textbf{\textit{High-order redundant feature masking:}}
\FOR{each node $v$}
    \FOR{$i=2$ to $K$}
        \FOR{each neighbor $u$ in $\mathbf{A}_i$}
            \STATE Calculate $\text{LCS}(v,u,i)$ by Eq.~\eqref{eq6}
        \ENDFOR
        \STATE Select top-$m_i$ neighbors by LCS, mask others in $\mathbf{A}_i$
    \ENDFOR
\ENDFOR
\STATE Update $\tilde{\mathbf{A}}$ and $\mathbf{H}_{\text{high}}$ using $\bar{\mathbf{A}}_i^{\text{filter}}$
\STATE \textbf{\textit{Feature fusion:}}
\STATE Learn $\beta \in [0,1]$
\STATE $\mathbf{H} = \beta \mathbf{H}_{\text{low}} + (1-\beta) \mathbf{H}_{\text{high}}$
\RETURN $\mathbf{H}$
\end{algorithmic}
\end{algorithm}